\documentclass[runningheads]{llncs}

\usepackage[final,year=2026,ID=9939]{eccv}

\usepackage{eccvabbrv}

\usepackage{graphicx}
\usepackage{booktabs}
\usepackage{arydshln}
\usepackage{float}
\usepackage[table]{xcolor}

\usepackage{wrapfig}
\usepackage{algorithm}
\usepackage{algorithmic}
\usepackage{arydshln}
\usepackage{xcolor}
\usepackage{amssymb}  
\usepackage{bm}
\usepackage{pifont}

\usepackage{multirow} 

\usepackage[accsupp]{axessibility}  

\usepackage{hyperref}

\usepackage{orcidlink}

\begin{document}

\title{GeoRect4D: Geometry-Compatible Generative Rectification for Dynamic Sparse-View 3D Reconstruction} 

\titlerunning{GeoRect4D for Dynamic Sparse-View 3D Reconstruction}


\author{Zhenlong Wu \and
Zihan Zheng \and
Xuanxuan Wang \and
Lei Huang \and
Hongwei Hu\and
Xiaoyun Zhang \and
Qiang Hu \and
Wenjun Zhang 
}

\authorrunning{Z. Wu et al.}

\institute{
Shanghai Jiao Tong University, Shanghai, China
}

\maketitle

\begin{figure}[H]
    \centering
    \vspace{-20pt}
    \includegraphics[width=\textwidth]{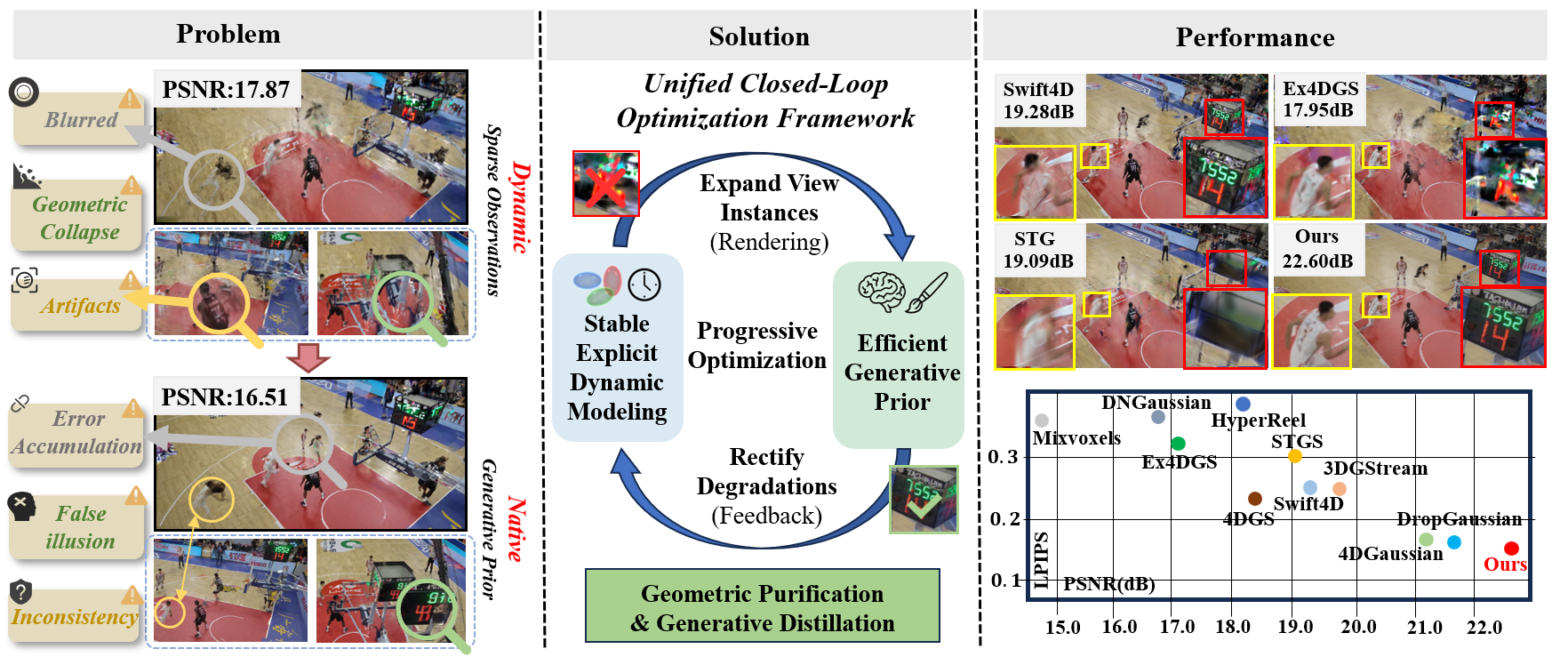} 
    \caption{
        \textbf{Left}: Sparse dynamic reconstruction suffers from geometric blur, while naive priors induce structural artifacts. \textbf{Middle}: Our closed-loop framework couples explicit dynamic modeling with a generative prior for progressive rectification. \textbf{Right}: GeoRect4D achieves superior visual fidelity and perceptual quality over other methods.
    }
    \label{fig:teaser}
    \vspace{-15pt}
\end{figure}

\begin{abstract}
Reconstructing dynamic 3D scenes from sparse multi-view videos is highly ill-posed, often leading to geometric collapse, trajectory drift, and floating artifacts. Recent attempts introduce generative priors to hallucinate missing content, yet naive integration frequently causes structural drift and temporal inconsistency due to the mismatch between stochastic 2D generation and deterministic 3D geometry. In this paper, we propose GeoRect4D, a novel unified framework for sparse-view dynamic reconstruction that couples explicit 3D consistency with generative refinement via a closed-loop optimization process. Specifically, GeoRect4D introduces a degradation-aware feedback mechanism that incorporates a robust anchor-based dynamic 3DGS substrate with a single-step diffusion rectifier to hallucinate high-fidelity details. This rectifier utilizes a structural locking mechanism and spatiotemporal coordinated attention, effectively preserving physical plausibility while restoring missing content. Furthermore, we present a progressive optimization strategy that employs stochastic geometric purification to eliminate floaters and generative distillation to infuse texture details into the explicit representation. Extensive experiments demonstrate that GeoRect4D achieves state-of-the-art performance in reconstruction fidelity, perceptual quality, and spatiotemporal  consistency across multiple datasets. Project Page: https://mediax-sjtu.github.io/GeoRect4D
\end{abstract}

\section{Introduction}
\label{sec:intro}

Reconstructing dynamic 3D scenes from sparse multi-view videos remains a fundamentally ill-posed problem. Without dense cross-view constraints, limited spatiotemporal cues cause optimization over explicit representations to converge to degenerate configurations, including geometric collapse, floaters, trajectory drift, and texture ambiguity. Such failures are more pronounced in dynamic scenes, where motion and geometry are tightly coupled under insufficient supervision. Solving this challenge is critical for practical 6-DoF free-viewpoint video (FVV) systems in resource-constrained capture settings.

Neural Radiance Fields (NeRF)~\cite{mildenhall2021nerf} and their dynamic extensions~\cite{park2021nerfies,pumarola2021d,cao2023hexplane,wang2022fourier} have demonstrated high-fidelity novel view synthesis under dense multi-view supervision, but typically incur substantial training and rendering costs. 
More recently, 3D Gaussian Splatting (3DGS)~\cite{kerbl20233d} offers an explicit and efficient alternative with real-time rendering and strong structural controllability, and has been extended to dynamic scenarios via deformation modeling and motion-aware parameterization~\cite{wu20244d,luiten2024dynamic,yang2024deformable,guo2024motion,zheng20254dgcpro}. 
Despite differences in representation and efficiency, most dynamic NeRF/3DGS methods rely on dense camera arrays to disambiguate geometry and motion. 
Under sparse inputs, optimization becomes unstable and generalization to novel views degrades.

To mitigate sparse-view ambiguity, prior work primarily focuses on constraining the solution space through additional regularization. 
NeRF-based approaches incorporate geometric supervision such as depth or normals~\cite{wang2023sparsenerf,yu2022monosdf} and frequency regularization~\cite{jain2021putting,yang2023freenerf}. 
Within the 3DGS framework, depth-guided constraints~\cite{li2024dngaussian,zhu2024fsgs} and dropout-based stabilization~\cite{park2025dropgaussian,xu2025dropoutgs} improve robustness under sparse inputs. 
While these techniques show promising results in static scenes, extending sparse reconstruction to dynamic scenarios remains significantly more challenging due to motion-induced ambiguity and temporal error accumulation. 
Existing sparse dynamic 3DGS methods~\cite{hu2025sparse4dgs,shi2025sparse4dgs} still struggle to maintain geometric stability under extremely limited supervision.

More recently, generative priors have been introduced to hallucinate missing content under sparse views. Several works leverage 2D generative models or diffusion-based refinement to enhance degraded renderings or augment pseudo-supervision~\cite{wu2024reconfusion,liu20243dgs,wu2025difix3d+}. 
However, most of these approaches are demonstrated on static scenes. In dynamic settings, naive integration of generative feedback often introduces new instabilities, as shown in Fig.~\ref{fig:teaser}. This failure stems from two fundamental barriers: significant geometric collapse within under-constrained representations hindering plausible hallucination, and the inherent mismatch between stochastic 2D generative refinement and the deterministic 3D geometric manifold. This conflict typically induces severe structural drift and spatiotemporal inconsistency, rendering the reconstruction unusable for immersive FVV. These observations suggest that sparse dynamic reconstruction is not merely a data scarcity problem, but a stability issue arising from interactions between under-constrained geometry and stochastic generative feedback. Our key insight is that to prevent structural drift, generative priors must be firmly anchored by explicit geometric constraints rather than serving as unconstrained 2D refinement.

To address these limitations, we propose GeoRect4D, a unified generative feedback framework for sparse-view dynamic reconstruction. Our key idea is to maintain a stable geometric substrate while progressively injecting generative knowledge through a closed-loop optimization process. Specifically, GeoRect4D couples an anchor-based dynamic 3DGS substrate with an efficient single-step diffusion rectifier to hallucinate missing details. The rectifier is fine-tuned to map degraded renderings to photorealistic imagery and employs structural locking together with spatiotemporal coordinated attention to preserve geometric consistency during generative refinement. 

To ensure stable convergence under sparse inputs, we further introduce a progressive optimization strategy. Instead of relying solely on soft regularization, our approach first performs stochastic geometric purification to remove floating artifacts and establish a stable geometric foundation. Subsequently, a generative distillation stage utilizes rectified renderings as high-quality pseudo-supervision, allowing the explicit 3D representation to gradually absorb high-frequency generative knowledge while maintaining physical plausibility. Extensive experiments demonstrate that GeoRect4D achieves state-of-the-art performance across multiple datasets. On the highly dynamic MPEG dataset, our method improves PSNR by up to \textbf{3.32 dB}, as illustrated in Fig.~\ref{fig:teaser}.

In summary, our key contributions include:

\begin{itemize}
\item We present GeoRect4D, a unified generative feedback framework for sparse-view dynamic reconstruction that stabilizes under-constrained geometry through closed-loop coupling between explicit 3D Gaussian representations and generative priors.
\item We introduce a generative rectification module based on a single-step diffusion prior together with a progressive optimization strategy, enabling consistent detail hallucination via structural locking, spatiotemporal attention, geometric purification, and generative distillation.
\item Extensive experimental results on multiple dynamic-scene datasets demonstrate that our GeoRect4D consistently outperforms existing methods in reconstruction fidelity, perceptual quality, and spatiotemporal  consistency.
\end{itemize}

\vspace{-10pt}

\section{Related Work}
\subsection{Dynamic Novel View Synthesis with NeRF}

NeRF\cite{mildenhall2021nerf} initially revolutionized 3D scene synthesis by modeling scenes as continuous implicit functions\cite{barron2021mip,barron2022mip,barron2023zip,chen2023dbarf,park2023camp,reiser2023merf}. Extending this paradigm to the temporal domain, early dynamic approaches typically employed a deformation field\cite{du2021neural,park2021nerfies,song2023nerfplayer,pumarola2021d,zheng2024hpc,zheng2024jointrf} to warp time-variant observations into a canonical space. However, the prohibitive training and rendering latencies of these implicit fields prompted the introduction of explicit structured representations\cite{cao2023hexplane,fridovich2023k,icsik2023humanrf,shao2023tensor4d,wang2023mixed,park2023temporal,li2022neural,wang2022fourier}. Despite these efficiency gains, these methods often incur high storage overheads and remain heavily reliant on dense multi-view supervision. Under sparse-view settings, they struggle to decouple complex motion from geometry, leading to severe blurring or topological artifacts.

\subsection{Dynamic Novel View Synthesis with 3DGS}
Recently, 3D Gaussian Splatting~\cite{kerbl20233d} has drawn significant attention for its high-fidelity reconstruction and real-time rendering, spurring numerous extensions~\cite{charatan2024pixelsplat,chen2024mvsplat,niedermayr2024compressed,radl2024stopthepop,yu2024mip,feng2025flashgs,hollein20253dgs}. To model dynamics, existing approaches fall into two paradigms. Deformation-based methods~\cite{qingming2025modgs,lu20243d,wu20244d,yang2024deformable,huang2024sc} warp canonical Gaussians via learnable fields into time-varying states. Though compact, they struggle with rapid topological changes. In contrast, explicit methods model temporal evolution through frame-wise tracking~\cite{guo2024motion,luiten2024dynamic,sun20243dgstream,gao2024hicom,hu20254dgc,zheng20254dgcpro} or unified 4D primitives~\cite{lee2024fully,li2025gifstream,wu2025swift4d,li2024spacetime,yang2023real}, achieving better resource efficiency and robustness to complex motion. However, both paradigms rely heavily on dense multi-view inputs to disentangle motion from geometry. Under sparse views, deformation fields suffer severe blurriness from under-constrained optimization, while explicit methods are prone to trajectory drift and artifacts, lacking the robust priors necessary to maintain temporal and geometric consistency.

\subsection{Sparse-View Novel View Synthesis}

Sparse-view novel view synthesis remains challenging due to severely under-constrained geometry and appearance. To mitigate this, NeRF and 3DGS methods often employ soft regularizations, including depth or normal priors~\cite{wang2023sparsenerf,yu2022monosdf,song2023darf,deng2022depth,li2024dngaussian,zhu2024fsgs,chung2024depth,xu2025depthsplat}, semantic or frequency regularization~\cite{jain2021putting,yang2023freenerf,niemeyer2022regnerf,truong2023sparf,qi2021multi}, and dropout-based stabilization~\cite{park2025dropgaussian,xu2025dropoutgs,zhang2024cor,han2024binocular,radl2024stopthepop}. However, these constraints still fail to prevent structural collapse or severe trajectory drift under extreme sparsity. Recently, diffusion models have emerged as powerful generative priors for novel view synthesis and 3D augmentation~\cite{gao2024cat3d,liu2023zero,wu2024reconfusion,liu2023deceptive,liu20243dgs,wu2025difix3d+}. Yet, naively applying them to dynamic scenes causes severe temporal flickering and structural drift, and existing methods lack explicit modeling of 3D reconstruction degradation and inherent spatiotemporal constraints. In contrast, our unified framework enables robust sparse dynamic reconstruction by explicitly modeling degradation and performing geometric purification.

\section{Method}

\begin{figure}[t]         
    \centering
    \includegraphics[width=\linewidth]{png/fig2.png} 
    \caption{Overview of GeoRect4D. \textbf{Left:} Sparse inputs are adaptively decomposed into a static base and an anchor-controlled dynamic field to construct the base dynamic 3DGS model. \textbf{Top Right:} A degradation-aware, single-step diffusion prior synthesizes high-fidelity rectified views from coarse renderings. \textbf{Bottom Right:} A two-stage progressive optimization framework first applies geometric purification to stabilize the substrate, then performs generative distillation to seamlessly integrate hallucinated details with strict physical fidelity.}
    \label{fig:architecture}
    \vspace{-8pt}
\end{figure}

\subsection{Problem Formulation and Framework Overview}
\label{sec:3.1}
We target high-fidelity dynamic 3D reconstruction from calibrated sparse multi-view videos
\begin{equation}
\mathcal{V} = \{(I_{t,v}, \pi_{t,v})\}_{t=1,v=1}^{T,N_v}, \qquad N_v \le 4,
\label{eq:data_sequence}
\end{equation}
where $I_{t,v}$ and $\pi_{t,v}$ denote the image and camera pose of view $v$ at time $t$.
Our goal is to recover a dynamic scene model that supports photorealistic novel-view synthesis at an arbitrary time and viewpoint $(t,\xi)$. Under such an extreme view budget, purely photometric supervision provides insufficient cross-view constraints, and direct optimization often collapses into degenerate geometry with floating artifacts and blurred appearance.
These failures are further amplified by temporal error accumulation in dynamic scenes.

To stabilize optimization in this under-constrained regime, we learn a unified rendering framework that couples an explicit geometric substrate $\mathcal{G}$  with a degradation-aware generative rectifier $\mathcal{D}$ in a closed loop (Fig.~\ref{fig:architecture}). The substrate is rendered by differentiable Gaussian splatting~\cite{kerbl20233d}.
At each timestamp $t$, the scene is represented as a set of Gaussians $\mathcal{G}(t)=\{g_i(t)\}$ with parameters
$(\boldsymbol{\mu}_i(t), \boldsymbol{\Sigma}_i(t), \alpha_i(t), \mathbf{f}_i(t))$ (center, covariance, opacity, and view-dependent appearance),
and rendering from viewpoint $\xi$ is
\begin{equation}
\hat{I}_{\mathrm{render}}(t,\xi) = \mathcal{R}(\mathcal{G}(t), \xi).
\label{eq:rendering}
\end{equation}

A key difficulty in sparse-view dynamics is that motion and geometry often become entangled early in training, which amplifies drift.
We therefore instantiate $\mathcal{G}$ with a stability-oriented disentangled parameterization: a static field $\mathcal{G}_s$
and a dynamic field $\mathcal{G}_d(t)$, i.e., $\mathcal{G}(t)=\mathcal{G}_s \cup \mathcal{G}_d(t)$.
Following an anchor-based dynamic formulation~\cite{lee2024fully}, we parameterize $\mathcal{G}_d(t)$ using temporal control anchors at nodal timestamps
$\mathcal{T}=\{t_0,\ldots,t_K\}$ and synthesize intermediate states via differentiable interpolation
\begin{equation}
\mathcal{G}_d(t) = \Phi\!\left(\mathcal{G}_d(t_k), \mathcal{G}_d(t_{k+1}), t\right), \qquad t\in[t_k,t_{k+1}].
\label{eq:anchor_interp}
\end{equation}
To make this formulation reliable under sparse supervision, we further adapt it by identifying dynamic primitives through accumulated positional-gradient statistics:
a Gaussian is assigned to $\mathcal{G}_d$ if its center-gradient magnitude exhibits outlier behavior
\begin{equation}
\nabla_{\mu}^{(i)} > \bar{\nabla}_{\mu} + \kappa \cdot \sigma_{\nabla},
\label{eq:dynamic_filtering}
\end{equation}
where $\bar{\nabla}_{\mu}$ and $\sigma_{\nabla}$ are computed over all primitives, and $\kappa$ controls the dynamic budget.
Intuitively, static primitives converge to stable positions, whereas truly dynamic ones keep receiving persistent gradients. 
This criterion prevents noisy motion assignment caused by sparse-view ambiguity.

Built on this stabilized substrate, we introduce a generative rectifier $\mathcal{D}$ to provide controlled feedback for sparse render degradations.
Given a rendering, the rectifier produces
\begin{equation}
\hat{I}_{\mathrm{rect}}(t,\xi) = \mathcal{D}\!\left(\hat{I}_{\mathrm{render}}(t,\xi)\right),
\label{eq:rectify}
\end{equation}
and we optimize $\mathcal{G}$ with a hybrid objective that couples physical observations and rectified supervision:
\begin{equation}
\min_{\mathcal{G}} \sum_{t,\xi}
\Big(
(1-\lambda(t))\,\mathcal{L}_{\mathrm{data}}(\hat{I}_{\mathrm{render}}, I_{t,v})
+\lambda(t)\,\mathcal{L}_{\mathrm{prior}}(\hat{I}_{\mathrm{render}}, \hat{I}_{\mathrm{rect}})
\Big).
\label{eq:joint_objective}
\end{equation}
Here $\mathcal{L}_{\mathrm{data}}$ anchors geometry to real observations, while $\mathcal{L}_{\mathrm{prior}}$ distills reliable details from the rectified views (Sec.~\ref{sec:3.2}).
Crucially, this disentangled architecture facilitates our progressive optimization strategy (Sec.~\ref{sec:3.3}). By first optimizing $\mathcal{G}_s$ and the anchor-based $\mathcal{G}_d(t)$ through a Geometric Purification phase, we establish a robust substrate that is physically consistent with the observed projections. This structure provides the necessary geometric foundation for the subsequent stage, where generative priors are applied to refine appearance details.

\subsection{Degradation-Aware Generative Prior}
\label{sec:3.2}
As introduced in Sec.~\ref{sec:3.1}, our rectifier $\mathcal{D}$ (Eq.~\ref{eq:rectify}) provides image pseudo-supervision for closed-loop optimization. 
Despite the stabilized geometric substrate, sparse-view supervision still leaves appearance severely under-constrained, resulting in hole-like missing regions, over-smoothed textures, and view-/time-dependent artifacts, especially around occlusions and thin structures. 
While a strong 2D diffusion prior could in principle restore plausible details, naively inserting an off-the-shelf model into our closed-loop pipeline is unsafe: 
(i) The VAE bottleneck may distort fine structures and deviate from the projected 3D geometry.
(ii) The degradation statistics of sparse 3D renderings differ substantially from the natural-image domain, and
(iii) frame-wise stochastic refinement introduces temporal flickering.
We thus instantiate $\mathcal{D}$ as a degradation-aware rectifier with targeted designs to ensure \textbf{geometry compatibility} and \textbf{temporal stability}.

\textbf{Structure-Preserving Feature Injection.}
We design $\mathcal{D}$ as an efficient single-step diffusion rectifier to enable iterative feedback without multi-step sampling overhead. To suppress structure drift caused by latent compression, we inject multi-scale spatial cues from the encoder into the decoder via projected skip connections. Let $\mathbf{h}_{enc}^{(l)}$ and $\mathbf{h}_{dec}^{(l)}$ denote encoder and decoder features at level $l$. The decoder update is
\begin{equation}
\mathbf{h}_{dec}^{(l)} \leftarrow 
\mathrm{UpBlock}\!\left(
\mathbf{h}_{dec}^{(l-1)} + C_{proj}^{(l)}\!\left(\gamma\,\mathbf{h}_{enc}^{(L-l)}\right)
\right),
\label{eq:diffusion_skip_final}
\end{equation}
where $C_{proj}^{(l)}$ is a learnable $1{\times}1$ projection aligning feature spaces and $\gamma$ controls the anchoring strength. This injection constrains refinement to remain aligned with the rendered 3D projection, suppressing diffusion-induced shape deviation.

\textbf{Degradation-Aware Residual Adaptation.}
To bridge the domain gap between natural-image priors and sparse 3D renderings, we adapt a frozen diffusion backbone via lightweight low-rank adaptation to predict artifact-specific residual corrections in latent space. Denoting the degraded latent by $\mathbf{z}_{deg}$ and the refined latent by $\mathbf{z}_{gt}$, we adopt a subtraction-based residual bridge:
\begin{equation}
\mathbf{z}_{gt} = \mathbf{z}_{deg} - \lambda_{res}\,\epsilon_{\theta}(\mathbf{z}_{deg}),
\label{eq:latent_residual_bridge_final}
\end{equation}
where $\epsilon_{\theta}$ is the adapted residual predictor and $\lambda_{res}$ controls refinement intensity. We set $\lambda_{res}=1$ during training and adjust it at inference to balance enhancement strength and structural fidelity. The rectifier is trained on paired samples rendered at training viewpoints across diverse 3D datasets (more details regarding the training datasets are available in the supplementary material), and regressing to the captured images using a pixel-perceptual objective:
\begin{equation}
\mathcal{L}_{\mathrm{rect}} =
\lambda_{pix}\|\hat{I}_{\mathrm{rect}} - I_{\text{gt}}\|_1
+ \lambda_{per}\,\mathrm{LPIPS}(\hat{I}_{\mathrm{rect}}, I_{\text{gt}}),
\label{eq:rect_loss_final}
\end{equation}
which jointly enforces low-level artifact suppression and perceptual consistency.

\textbf{Spatio-Temporal Coordinated Attention.}
Temporal instability becomes particularly critical when rectified views are distilled back into 3D optimization. To reduce frame-wise randomness, we refine the current latent jointly with coordinated spatial and temporal references. For frame $t$, we construct a reference set $\mathcal{W}_{ref}=\{\mathbf{z}_{spa},\mathbf{z}_{tem}\}$, where $\mathbf{z}_{spa}$ aggregates latents from other available views at the same timestamp and $\mathbf{z}_{tem}$ aggregates a short temporal window around $t$. The coordinated refinement is expressed as
\begin{equation}
\mathbf{z}_{out} =
\mathrm{STCAttn}(\mathbf{z}_t;\mathcal{W}_{ref})
= \mathcal{T}^{-1}\!\Big(
\mathrm{SelfAttn}\big(\mathcal{T}([\mathbf{z}_t,\mathcal{W}_{ref}])\big)
\Big),
\label{eq:stc_attn_final}
\end{equation}
where $\mathcal{T}:\mathbb{R}^{V\times N\times C}\rightarrow\mathbb{R}^{(V\cdot N)\times C}$ flattens view and spatial-token dimensions for global interaction. This spatio-temporal coordination suppresses flickering and yields stable rectified targets for subsequent closed-loop distillation.

\subsection{Progressive Geometric Purification and Generative Distillation
}
\label{sec:3.3}

\begin{figure}[t]          
    \centering
    \includegraphics[width=\linewidth]{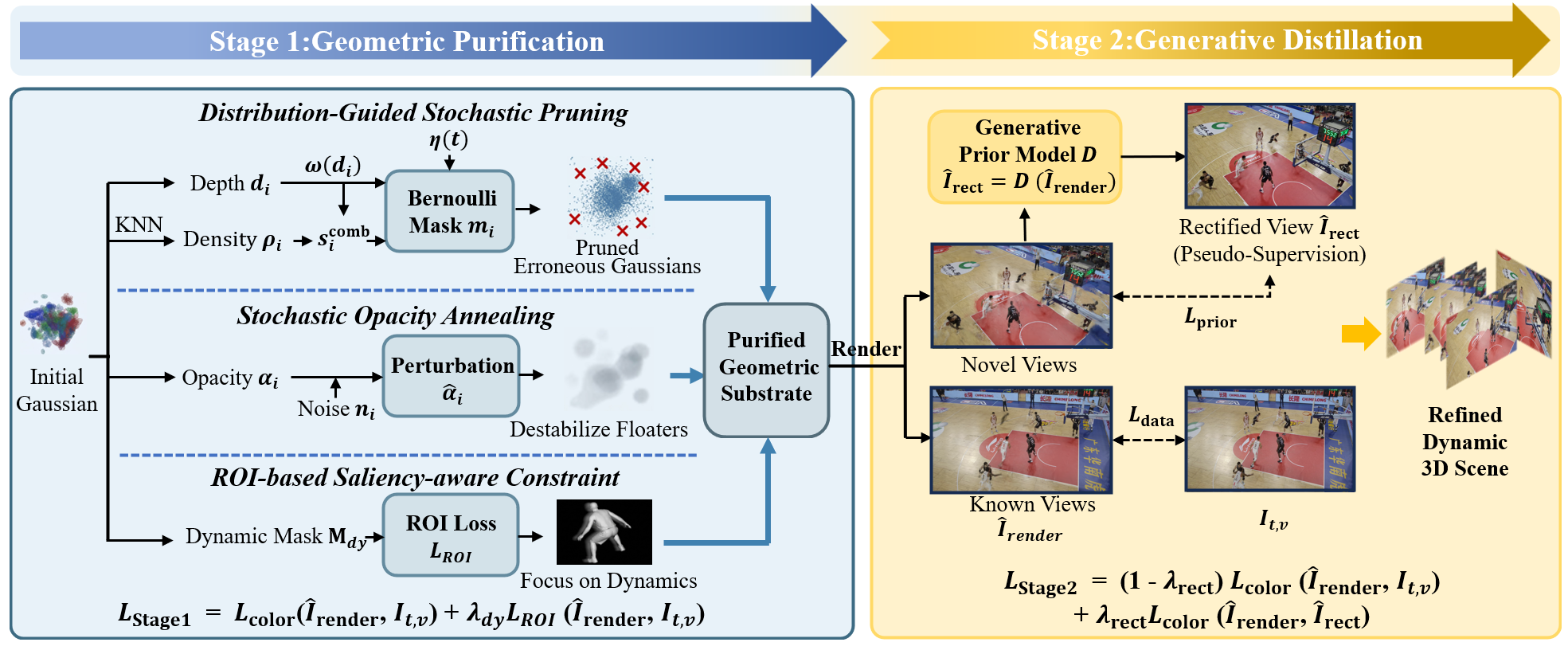} 
    \caption{Illustration of our two-stage progressive optimization framework. Stage 1 executes geometric purification via stochastic pruning, opacity annealing, and ROI-based dynamic constraints to establish a robust substrate. Stage 2 performs generative distillation, transferring details from the generative prior into the 3D scene through a hybrid objective combining physical and pseudo-supervision.}
    \label{fig:progressive}
    \vspace{-8pt}
\end{figure}

We establish a progressive optimization framework to bridge the substantial gap between sparse physical observations and photorealistic rendering. Diverging from standard pipelines that often compromise between geometric accuracy and textural detail, our approach orchestrates a two-phase evolution: first executing \textbf{Geometric Purification} to eliminate artifacts from the substrate, and subsequently performing \textbf{Generative Distillation} to transfer generative priors into the 3D representation.

\textbf{Stage 1: Stochastic Geometric Purification:} 
We observe that sparse supervision often overfits static scenes, minimizing projection errors by generating low-depth, high-density anomaly clusters rather than physically plausible geometry. Since classical hard pruning may prematurely remove unconverged valid structures, we propose a \textbf{Distribution-Guided Stochastic Pruning}. Specifically, for each Gaussian $i$, we compute its depth $d_i$ relative to the camera center and estimate its local spatial density $\rho_i$ via the K-Nearest Neighbors (KNN) algorithm. We then apply a time-varying dropout pressure $\eta(t)$ based on the training progress. The binary existence mask $m_i$ for the $i$-th Gaussian is formulated as a random variable following a Bernoulli distribution:
\begin{equation}
    m_i \sim \mathrm{Bernoulli}(p_i), \quad \text{where } p_i = 1 - \eta(t) \cdot \omega(d_i) \cdot s_i^{\text{comb}}
    \label{eq:bernoulli_mask}
\end{equation}
where $\eta(t)$ increases linearly during training; $\omega(d_i)$ is a continuous penalty weight decaying linearly with depth; and $s_i^{\text{comb}}$ represents the min-max normalized weighted sum of the inverse depth $1/d_i$ and the density $\rho_i$. Acting as a probabilistic high-pass filter, this soft Bernoulli process gradually eliminates dense, near-camera erroneous floaters without prematurely destroying valid physical geometry.

Simultaneously, to suppress shape-radiance ambiguity caused by sparse observations, which often leads the optimizer to minimize photometric loss by stacking semi-transparent floaters, we introduce a \textbf{Stochastic Opacity Annealing} strategy. We inject multiplicative noise into opacities via $\hat{\alpha}_i = \mathrm{clamp}(\alpha_i \cdot (1 + \mathbf{n}_i), 0, 1)$ with $\mathbf{n}_i \sim \mathcal{N}(0, \sigma^2(t))$, where the standard deviation $\sigma(t)$ decays linearly from $0.3$ to $0$ as training progresses. This continuous perturbation acts as a geometric stress test: it amplifies the rendering variance of fragile, semi-transparent floaters, destabilizing their gradient flow and compelling the model to seek geometric configurations that remain stable under stochastic noise, without incurring inference overhead.

Furthermore, under sparse supervision, dominant errors from static regions often overwhelm dynamic details, causing motion blurring and ghosting. To address this, we introduce an\textbf{ ROI-based Saliency-aware Constraint}. Specifically, we extract a 2D dynamic mask $\mathbf{M}_{dy}$ by exclusively rendering all dynamic Gaussians and systematically filtering out spatial outliers. Leveraging $\mathbf{M}_{dy}$ to guide the data fidelity term $\mathcal{L}_{\text{data}}$, and setting the generative coupling factor $\lambda = 0$ (Eq.~\ref{eq:joint_objective}) to focus entirely on robust geometric initialization, the Stage 1 optimization objective is formulated as:
\begin{equation}
\begin{split}
    \mathcal{L}_{\text{Stage1}} =\mathcal{L}_{\text{data}}= \underbrace{(1 - \lambda_{s}) \mathcal{L}_1 + \lambda_{s} \mathcal{L}_{\text{SSIM}} }_{\mathcal{L}_{\text{color}}(\hat{I}_{\text{render}}, I_{t,v})}
    + \lambda_{dy} \underbrace{\| \mathbf{M}_{dy} \odot (\hat{I}_{\text{render}} - I_{t,v}) \|_1}_{\mathcal{L}_{ROI}(\hat{I}_{\text{render}}, I_{t,v})}
    \label{eq:roi_loss}
\end{split}
\end{equation}
where  $ \odot $  denotes the Hadamard product. This term encourages the optimizer to place greater emphasis on sparse dynamic pixels, ensuring that fine-grained motion details are recovered rather than diluted by static field optimization.

\textbf{Stage 2: Generative Knowledge Distillation.} 
Upon establishing a purified geometric substrate, we address the inherent information deficit of extreme view sparsity. We sample unobserved novel viewpoints and unlock the potential of our diffusion prior (Sec.~\ref{sec:3.2}) to enhance their initial renderings into high-fidelity rectified views $\hat{I}_{\text{rect}}$. Crucially, while the known training views remain strictly supervised by ground truth, we treat these rectified novel views as a source of pseudo-supervision to transfer generative knowledge back into the explicit 3D representation. Specifically, we activate the generative feedback loop by instantiating the coupling factor $\lambda(t)$ (Eq.~\ref{eq:joint_objective}) as a non-zero weight $\lambda_{\text{rect}}$. This weight serves as a confidence regulator, preventing potential geometric distortion caused by over-reliance on hallucinated content. To construct the hybrid feedback objective, both the data fidelity term $\mathcal{L}_{\text{data}}$ (anchored by physical observations) and the prior term $\mathcal{L}_{\text{prior}}$ (guided by pseudo-supervision) in Eq.~\ref{eq:joint_objective} are instantiated with the photometric loss $\mathcal{L}_{\text{color}}$:
\begin{equation}
    \mathcal{L}_{\text{Stage2}} = (1 - \lambda_{\text{rect}}) \mathcal{L}_{\text{color}}(\hat{I}_{\text{render}}, I_{t,v}) + \lambda_{\text{rect}} \mathcal{L}_{\text{color}}(\hat{I}_{\text{render}}, \hat{I}_{\text{rect}})
    \label{eq:feedback_loss}
\end{equation}
where $ \lambda_{\text{rect}} $  is the weighting factor for the rectified signal, typically set to a small value, ensuring that the optimization remains grounded in physical observations $I_{t,v}$ while selectively absorbing high-frequency textures from the rectified views $\hat{I}_{\text{rect}}$. This strategy effectively distills the strong generative power of 2D diffusion models into the explicit 3D scene representation, enabling knowledge transfer from sparse input to dense details.

\section{Experiments}

\begin{table}[!htbp]
  \caption{Quantitative comparison on the N3DV\cite{li2022neural} and MeetRoom\cite{li2022streaming} datasets. The reported metrics are averaged over all 300 frames for each sequence. \colorbox{red!25}{Red}, \colorbox{orange!25}{orange}, and \colorbox{yellow!25}{yellow} backgrounds indicate the best, second-best, and third-best results, respectively. $^*$STGS was trained in sub-groups to complete the 300-frame sequences due to GPU memory constraints.}
  \label{tab:main_results}
  \centering
  \resizebox{\textwidth}{!}{

  \setlength{\tabcolsep}{3pt} 
  \begin{tabular}{l ccccc ccccc}
    \toprule
    & \multicolumn{5}{c}{N3DV Dataset} & \multicolumn{5}{c}{MeetRoom Dataset} \\
    \cmidrule(lr){2-6} \cmidrule(l){7-11}
    Method & PSNR$\uparrow$ & SSIM$\uparrow$ & LPIPS$\downarrow$ & tOF$\downarrow$ & MUSIQ$\uparrow$  
    & PSNR$\uparrow$ & SSIM$\uparrow$ & LPIPS$\downarrow$ & tOF$\downarrow$ & MUSIQ$\uparrow$  \\
    \midrule
    HyperReel\cite{attal2023hyperreel} & 19.51 & 0.722 & 0.399 & 2.568 & 41.47 & 17.37 & 0.658 & 0.441 & 0.810 & 41.40 \\
    Mixvoxels\cite{wang2023mixed}      & 15.13 & 0.608 & 0.552 & 1.224 & 40.25 & 12.52 & 0.474 & 0.671 & 0.535 & 41.09 \\
    4DGaussian\cite{wu20244d}          & \cellcolor{yellow!25}24.97 & \cellcolor{yellow!25}0.857 & 0.225 & \cellcolor{yellow!25}1.022 & 57.26 & 12.20 & 0.483 & 0.611 & 0.576 & 44.49   \\
    STGS$^*$\cite{li2024spacetime}     & 19.99 & 0.732 & 0.457 & 1.219 & 46.09 & \cellcolor{orange!25}21.39 & \cellcolor{yellow!25}0.822 & \cellcolor{orange!25}0.157 & 0.494 & 45.17  \\
    4DGS\cite{yang2023real}            & 21.74 & 0.756 & 0.202 & \cellcolor{orange!25}1.008 & \cellcolor{yellow!25}58.70 & 18.15 & 0.713 & 0.295 & \cellcolor{orange!25}0.450 & \cellcolor{yellow!25}57.04  \\
    3DGStream\cite{sun20243dgstream}   & 18.54 & 0.685 & 0.422 & 1.287 & 48.90 & 13.44 & 0.506 & 0.569 & 0.634 & 51.69  \\
    Swift4D\cite{wu2025swift4d}        & \cellcolor{orange!25}25.99 & \cellcolor{orange!25}0.880 & \cellcolor{orange!25}0.144 & 1.064 & \cellcolor{orange!25}59.36 & \cellcolor{yellow!25}21.38 & \cellcolor{orange!25}0.840 & \cellcolor{yellow!25}0.162 & \cellcolor{yellow!25}0.455 & \cellcolor{orange!25}63.40 \\
    Ex4DGS\cite{lee2024fully}          & 24.28 & 0.848 & \cellcolor{yellow!25}0.184 & 1.061 & 52.76 & 18.36 & 0.800 & 0.248 & 0.488 & 38.88  \\
    Sparse4DGS\cite{hu2025sparse4dgs}  & 22.22 & 0.797 & 0.288 & - & - & - & - & - & - & - \\
    \midrule
    \textbf{Ours} & \cellcolor{red!25}26.98 & \cellcolor{red!25}0.884 & \cellcolor{red!25}0.113 & \cellcolor{red!25}0.994 & \cellcolor{red!25}62.73
                  & \cellcolor{red!25}22.23 & \cellcolor{red!25}0.878 & \cellcolor{red!25}0.120 & \cellcolor{red!25}0.432 & \cellcolor{red!25}68.20 \\
    \bottomrule
  \end{tabular}
  }
\vspace{-1em}
\end{table}

\subsection{Experimental Settings}

\textbf{Datasets.} We evaluate GeoRect4D on three real-world datasets: N3DV~\cite{li2022neural}, MeetRoom~\cite{li2022streaming}, and the high-motion MPEG dataset. For N3DV and MeetRoom, we use 300 frames at half resolution, selecting cameras 1, 4, 7, and 10 as sparse 4-view inputs. On the challenging MPEG dataset, we conduct experiments on the first 60 frames, utilizing cameras 0, 7, 14, and 21 for training. Following standard protocols, all models are evaluated on a single held-out camera (camera 0 for N3DV/MeetRoom and camera 8 for MPEG) as the novel view across all frames.

\textbf{Evaluation Metrics.} We report the PSNR and SSIM \cite{wang2004image} of all methods to evaluate their reconstruction quality and modeling capability, alongside the average LPIPS \cite{zhang2018unreasonable} for perceptual quality. For the temporal consistency evaluation of the reconstructed sequences, we adopt tLPIPS\cite{chu2020learning} and tOF \cite{chu2020learning}, and use RAFT \cite{teed2020raft} to compute optical flow. We also report no-reference metrics such as MUSIQ \cite{ke2021musiq} and MANIQA \cite{yang2022maniqa} as references to evaluate the visual quality of the enhanced images.

\textbf{Implementation Details.} Our framework is evaluated on a single NVIDIA RTX 3090 GPU. We build the generative prior upon the SD-Turbo backbone, applying trainable LoRA modules to all convolutional and linear layers, with the adapter loss weights set to $\lambda_{pix}=1.0$ and $\lambda_{per}=1.0$. Our progressive training is optimized for 20000 iterations per stage. Across all sequences, we set the threshold factor $\kappa=3.0$, the SSIM loss weight $\lambda_{s}=0.2$, and the dynamic weight $\lambda_{dy}=0.2$. During the generative feedback phase, we sample novel views via camera pose interpolation and utilize the diffusion prior for joint optimization, with the coupling factor set to $\lambda_{rect}=0.1$. More details regarding the camera pose sampling strategy are provided in the supplementary material.

\subsection{Comparison}

\begin{table}[!h]
  \caption{Quantitative comparison on the challenging MPEG dataset. The reported metrics are averaged over the first 60 frames. \colorbox{red!25}{Red} and \colorbox{orange!25}{orange} backgrounds indicate the best and second-best results, respectively.}
  \label{tab:cba_results}
  \centering
  \resizebox{\textwidth}{!}{
  \setlength{\tabcolsep}{5.5pt} 
  \begin{tabular}{l ccc cc cc}
    \toprule
    & \multicolumn{3}{c}{Fidelity} 
    & \multicolumn{2}{c}{Consistency} 
    & \multicolumn{2}{c}{Quality} \\
    
    \cmidrule(lr){2-4} \cmidrule(lr){5-6} \cmidrule(lr){7-8} 
    
    Method& PSNR$\uparrow$ & SSIM$\uparrow$ & LPIPS$\downarrow$ 
    & tOF$\downarrow$ & tLPIPS$\downarrow$ 
    & MUSIQ$\uparrow$ & MANIQA$\uparrow$ \\
    \midrule

    DNGaussian\cite{li2024dngaussian}  & 17.09 & 0.488 & 0.429 & 114.311 & 384.53 & 54.45 & 0.2849 \\
    DropGaussian\cite{park2025dropgaussian}      & \cellcolor{orange!25}21.65 & \cellcolor{red!25}0.789 & \cellcolor{orange!25}0.179 & 15.403 & 106.24 & \cellcolor{orange!25}62.11 & \cellcolor{orange!25}0.2984 \\
    3DGStream\cite{sun20243dgstream}        & 19.70 & 0.650 & 0.266 & 2.013 & 6.65 & 57.66 & 0.2721 \\
    STGS\cite{li2024spacetime}               & 19.09 & 0.688 & 0.300 & 1.799 & 18.47 & 45.13 & 0.2483 \\
    Ex4DGS\cite{lee2024fully}               & 17.95 & 0.624 & 0.316 & \cellcolor{orange!25}1.660 & 7.88 & 39.01 & 0.1482 \\
    Swift4D\cite{wu2025swift4d}             & 19.28 & 0.647 & 0.267 & 1.731 & \cellcolor{orange!25}4.35 & 52.07 & 0.2511 \\
    \midrule
    \textbf{Ours}                           & \cellcolor{red!25}22.60 & \cellcolor{orange!25}0.744 & \cellcolor{red!25}0.175 & \cellcolor{red!25}1.412 & \cellcolor{red!25}4.21  & \cellcolor{red!25}62.71 & \cellcolor{red!25}0.2987 \\
    \bottomrule
  \end{tabular}

  }
\vspace{-8pt}
\end{table}

\textbf{Quantitative Comparisons.} To validate the effectiveness of our method, we first compare it against state-of-the-art NeRF- and 3DGS-based dynamic view synthesis approaches including HyperReel\cite{attal2023hyperreel}, Mixvoxels\cite{wang2023mixed},  STGS\cite{li2024spacetime} , 4DGS\cite{yang2023real}, Swift4D\cite{wu2025swift4d}, 4DGaussian\cite{wu20244d}, Ex4DGS\cite{lee2024fully} and Sparse4DGS~\cite{hu2025sparse4dgs} on the N3DV and MeetRoom datasets. As shown in Tab. \ref{tab:main_results}, GeoRect4D consistently achieves the highest reconstruction quality across both datasets. While Sparse4DGS improves geometric stability via spatiotemporal regularization, it inherently lacks the generative capacity to hallucinate high-frequency details in under-observed regions. In contrast, our generative feedback framework overcomes this limitation, yielding a striking PSNR improvement (\textbf{26.98 dB} vs. 22.28 dB) alongside a drastic LPIPS reduction (\textbf{0.113} vs. 0.288) on N3DV. Beyond spatial fidelity, our method effectively mitigates trajectory drift and floating artifacts. This robust performance is further evidenced by highly competitive temporal consistency (a tOF of \textbf{0.994} on N3DV) and superior no-reference visual quality, securing the highest MUSIQ scores of \textbf{62.73} and \textbf{68.20} on the respective datasets.

To further evaluate the robustness of our method under extreme large-motion and topological changes, we conduct comparisons on the highly challenging MPEG dataset (Tab. \ref{tab:cba_results}). In this regime, the limitations of existing paradigms become glaringly apparent. Conventional dynamic baselines (e.g., Swift4D at 19.28 dB, Ex4DGS at 17.95 dB) suffer from severe motion-induced ambiguity and structural collapse. Conversely, while applying static sparse priors frame-by-frame (e.g., DropGaussian~\cite{park2025dropgaussian} and DNGaussian~\cite{li2024dngaussian}) can preserve some spatial structures, they completely fail to maintain temporal coherence, resulting in severe flickering and a massive tOF error of 15.403.  In stark contrast, our closed-loop generative feedback thrives in highly dynamic scenarios. GeoRect4D not only achieves a commanding PSNR of \textbf{22.60 dB} but also ensures that the hallucinated details remain structurally stable across time. By achieving the lowest tOF (\textbf{1.412}) and tLPIPS (\textbf{4.21}) alongside state-of-the-art perceptual realism (MUSIQ: \textbf{62.71}, MANIQA: \textbf{0.2987}), we demonstrate that our method fundamentally resolves the tension between spatial hallucination and temporal consistency in large-motion scenes.

\begin{figure}[t]   
\vspace{-3pt}
    \centering
    \includegraphics[width=\linewidth]{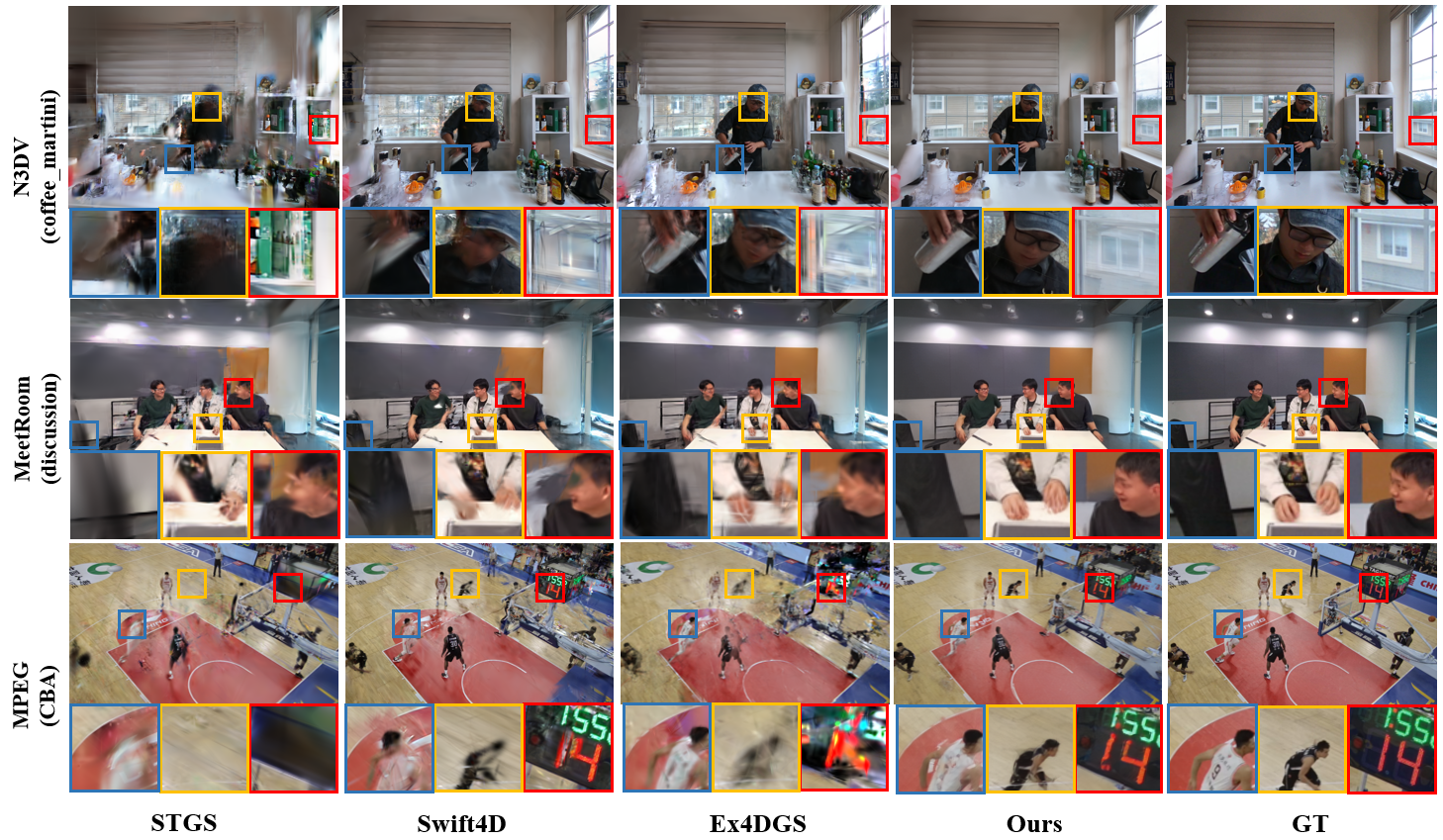} 
    \caption{Qualitative comparison of GeoRect4D with STGS\cite{li2024spacetime}, Swift4D\cite{wu2025swift4d} , and Ex4DGS\cite{lee2024fully}  across the N3DV\cite{li2022neural}, MeetRoom\cite{li2022streaming}, and MPEG datasets. 
    }
    \label{fig:comparsion}
    \vspace{-13pt}
\end{figure}

\textbf{Qualitative Comparisons.} We present qualitative comparisons with STGS, Swift4D, and Ex4DGS on the \textit{coffee\_martini}, \textit{discussion}, and \textit{CBA} sequences, as shown in Fig.~\ref{fig:comparsion}. Under the severely under-constrained 4-view setting, existing methods struggle to disentangle complex motion from geometry. Consequently, STGS suffers from geometric collapse, while Swift4D and Ex4DGS exhibit pervasive blurring and ghosting. Compared to these baselines, our GeoRect4D more effectively preserves finer structural boundaries and high-frequency details. Specifically, our method successfully recovers sharp player silhouettes and clear scoreboard digits in \textit{CBA}, as well as distinct facial expressions and fast-moving hands in the indoor sequences, which are heavily distorted or entirely lost in the baseline reconstructions. This demonstrates that our approach effectively suppresses erroneous floaters and hallucinates high-fidelity details, achieving a level of perceptual realism that closely matches the ground truth.

\subsection{Evaluations}

To rigorously evaluate the contribution of each proposed component, we conduct extensive ablation studies on the challenging MPEG dataset. Specifically, we systematically investigate the efficacy of our generative prior, the geometric purification and the generative rectification module. Furthermore, we provide an analysis of the computational efficiency of our framework.

\begin{table}[!htbp]
  \caption{Ablation study of the Degradation-Aware Generative Prior. \textbf{Struct.}: Structure-Preserving Module; \textbf{STCAttn}: Spatio-Temporal Coordinated Attention; \textbf{DA Adapter}: Degradation-Aware Adapter.}
  \label{tab:component_ablation}
  \centering
  \resizebox{\textwidth}{!}{

  \setlength{\tabcolsep}{5pt} 
  \begin{tabular}{ccc cc cc cc}
    \toprule
    \multicolumn{3}{c}{Components} 
    & \multicolumn{2}{c}{Fidelity} 
    & \multicolumn{2}{c}{Quality} 
    & \multicolumn{2}{c}{Consistency} \\
    
    \cmidrule(lr){1-3} \cmidrule(lr){4-5} \cmidrule(lr){6-7} \cmidrule(lr){8-9}
    
    Struct. & STCAttn & DA Adapter 
    & PSNR$\uparrow$ & SSIM$\uparrow$ 
    & MUSIQ$\uparrow$ & MANIQA$\uparrow$ 
    & tOF$\downarrow$ & tLPIPS$\downarrow$ \\
    \midrule
    
    \checkmark & \checkmark  & $\times$
    & 21.04 & 0.721 & 53.57 & 0.2444 & 1.535 & 5.23 \\
    
    $\times$ & $\times$ & \checkmark 
    & 21.98 & 0.723 & 43.21 & 0.1938 & 1.618 & \cellcolor{orange!25}4.71 \\
    
    \checkmark & $\times$ & \checkmark 
    & \cellcolor{orange!25}22.31 & \cellcolor{orange!25}0.738 & \cellcolor{red!25}63.63 & \cellcolor{orange!25}0.2943 & \cellcolor{orange!25}1.462 & 5.55 \\
    
    \checkmark & \checkmark & \checkmark 
    & \cellcolor{red!25}22.60 & \cellcolor{red!25}0.744 & \cellcolor{orange!25}62.71 & \cellcolor{red!25}0.2987 & \cellcolor{red!25}1.412 & \cellcolor{red!25}4.21 \\
    \bottomrule
  \end{tabular}

  }
  \vspace{-5pt}
\end{table}

\begin{figure}[t]         
    \centering
    \includegraphics[width=\linewidth]{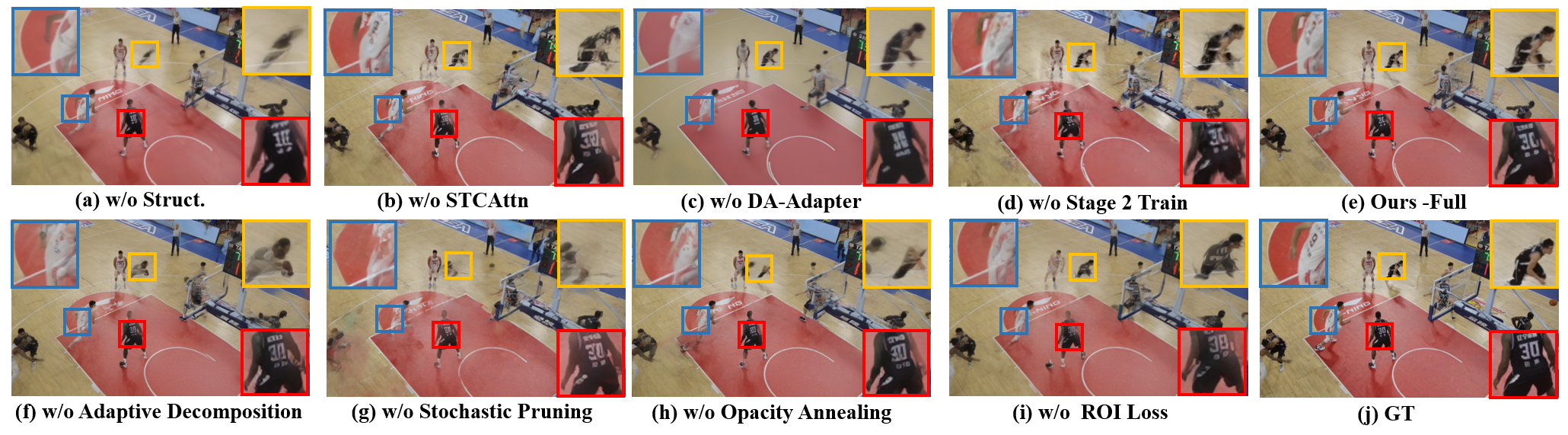} 
    \caption{Qualitative results of GeoRect4D and its variants on the MPEG dataset. Excluding any module leads to  severe blurring and structural artifacts.}
    \vspace{-1em}
    \label{fig:Ablation}
    \vspace{-5pt}
\end{figure}

\textbf{Generative Prior.} We first dissect the generative prior module introduced in Sec.~\ref{sec:3.2} to evaluate the impact of its specific designs. As detailed quantitatively in Tab. \ref{tab:component_ablation}, disabling these key components leads to significant performance drops across both perceptual and temporal metrics. Specifically, disabling the structure-preserving module leads to severe structural drift, causing the generative model to hallucinate implausible shapes and visible anatomical distortions on the player, which deteriorates the MANIQA score from \textbf{0.2987} to \textbf{0.1938}. Furthermore, applying 2D priors purely frame-by-frame by removing the spatio-temporal coordinated attention inherently compromises sequence consistency, increasing the tOF error from \textbf{1.412} to \textbf{1.462}.
Consistent with these numerical declines, Fig. \ref{fig:Ablation} illustrates the corresponding visual degradation. Without the degradation-aware adapter, the model fails to rectify specific artifacts, yielding noticeably blurry textures. More critically, removing the structure-preserving module detaches the synthesis from the underlying 3D geometry, causing severe structural drift and visible anatomical distortions on the player. Finally, omitting temporal attention leads to pronounced temporal jitter, highlighting its necessity for cross-frame stability.

\textbf{Geometric Purification and Generative Rectification.} As detailed in Tab. \ref{tab:ablation}, we systematically validate our geometric purification and generative rectification components. Within the purification stage, omitting adaptive decomposition yields a \textbf{0.53dB} PSNR drop, while discarding stochastic pruning leads to a catastrophic \textbf{1.36dB} degradation. Furthermore, opacity annealing is vital for suppressing shape-radiance ambiguities; without this perturbation, the optimizer overfits sparse views with semi-transparent floaters, dropping the SSIM to \textbf{0.743}. Similarly, removing the ROI loss dilutes the focus on dynamic regions, reducing the SSIM by \textbf{0.02}. Finally, completely removing the closed-loop generative rectification drastically drops the overall PSNR from \textbf{22.60dB} to \textbf{20.54dB}, and significantly worsens the perceptual LPIPS score from \textbf{0.175} to \textbf{0.224}. As visually confirmed by the qualitative degradations in Fig. \ref{fig:Ablation}, this massive gap definitively underscores that explicit 3D modeling alone is fundamentally insufficient to resolve extreme sparse-view ambiguities without our generative feedback.

\begin{table}[!t]
\vspace{-3pt}
  \caption{Ablation study of the Geometric Purification components and the overall Generative Rectification framework. `w/o' denotes without.}
  \vspace{-1em}
  \label{tab:ablation}
  \centering
  
  \resizebox{0.95\textwidth}{!}{

  \setlength{\tabcolsep}{9pt} 
  \begin{tabular}{l cccc}
    \toprule
    Setup & PSNR$\uparrow$ & SSIM$\uparrow$ & LPIPS$\downarrow$ & Time(min)$\downarrow$ \\
    \midrule
    w/o Adaptive Decomposition & 22.07 & 0.739 & 0.179 & 48.94 \\
    w/o Stochastic Pruning    & 21.24 & 0.724 & 0.194 & 50.84 \\
    w/o Opacity Annealing      & 22.34 & \cellcolor{orange!25}0.743 & 0.179 & 48.30 \\
    w/o ROI Loss               & \cellcolor{orange!25}22.44 & 0.742 & \cellcolor{red!25}0.175 & \cellcolor{orange!25}47.86 \\
    w/o Generative Rectification  & 20.54 & 0.694 & 0.224 & \cellcolor{red!25}42.57 \\ 
    \midrule
    \textbf{Ours-Full} & \cellcolor{red!25}22.60 & \cellcolor{red!25}0.744 & \cellcolor{red!25}0.175 & 49.96 \\
    \bottomrule
  \end{tabular}
  }
\end{table}

\begin{table}[t]
\centering
\caption{Computational efficiency breakdown of GeoRect4D across the evaluated datasets. We report the training duration for both progressive optimization stages, the generative rectification latency, and the final real-time rendering speed.}
\vspace{-1em}
\label{tab:efficiency_breakdown}

\resizebox{\linewidth}{!}{
\setlength{\tabcolsep}{6.5pt} 

\begin{tabular}{lcccc} %
\toprule

\textbf{Dataset} & Stage 1(min) & Stage 2(min) & Rectify Latency(ms) & Render Speed(FPS) \\
\midrule
N3DV       & 49.45 & 58.57 & 882 & 115 \\
MeetRoom   & 22.45 & 29.12 & 357 & 130 \\
MPEG       & 20.40 & 26.85 & 542 & 121 \\
\bottomrule
\end{tabular}
}
\vspace{-1em}
\end{table}

\textbf{Efficiency Analysis.} As detailed in Tab. \ref{tab:efficiency_breakdown}, while our closed-loop framework introduces additional training steps, the single-step nature of our diffusion prior minimizes generative rectification latency to under \textbf{882ms}. This efficiency enables our framework to achieve a highly competitive training time on the complex MPEG dataset, while maintaining real-time rendering speeds exceeding \textbf{115FPS}. More details on timing and resource consumption are available in the supplementary material.

\section{Discussion}

\textbf{Conclusion.} This paper presents GeoRect4D designed to tackle the highly ill-posed problem of dynamic 3D scene reconstruction from extremely sparse multi-view videos. By synergistically coupling a robust explicit dynamic 3D Gaussian Splatting substrate with a degradation-aware generative prior, our approach effectively overcomes prevalent issues such as geometric collapse, trajectory drift, and floating artifacts. Extensive evaluations demonstrate that GeoRect4D establishes a new state-of-the-art in both rendering fidelity and perceptual realism. By successfully hallucinating fine-grained details while strictly preserving spatiotemporal consistency, this work bridges the gap between physical geometric constraints and 2D generative priors for high-quality free-viewpoint video synthesis.

\textbf{Limitations.} Despite its effectiveness, GeoRect4D has two primary limitations. First, the framework's performance remains sensitive to SfM-based initialization, as textureless regions or extreme occlusions can result in suboptimal reconstruction. Second, although the single-step diffusion prior is efficient, the closed-loop generative distillation still introduces higher training overhead compared to purely explicit dynamic modeling.



%
%
\bibliographystyle{splncs04}
\bibliography{main}
\end{document}